# DriveWorld3D

## *Evidence Constrained 3D World Modeling and Generative Simulation for Real-World Autonomous Driving*

Robostreet research team

Cambridge, MA 02142 USA — Robostreet Inc.

Wei Wang*, Yiru Veronika Wang, Yihan Wang, Zihao Li, Xiaohui Liang

*Corresponding author, Email: weiwang9@mit.edu

Research status: This paper proposes a system architecture, implementable algorithmic definitions, and an experimental protocol intended for preregistration. The system has not yet been implemented or validated on a vehicle. No performance improvements are reported. The work does not reproduce a production driving system or establish a safety certification.

## Abstract

Three dimensional modeling for autonomous driving must continuously estimate road geometry, obstacle occupancy, traffic participant behavior, and observational uncertainty under limited computational resources. Existing reconstruction methods can synthesize realistic novel views, but visual fidelity does not sufficiently characterize the geometric reliability required for driving. Generative world models can expand scenarios and predict future observations, yet may introduce content unsupported by measurements. We propose DriveWorld3D, a framework combining temporal video learning, measurement constraints from multiple sensors, and offline generative simulation. Its online component maintains a unified world state comprising road surfaces, volumetric occupancy, dynamic entities, and road relations, while separately representing physical occupancy, observability, and evidence provenance. Action conditioned future prediction and risk sensitive planning connect this representation to driving decisions. Its offline component reconstructs recorded scenes using neural radiance fields or 3D Gaussian Splatting, then expands long tail conditions using generative models while restricting the use of generated content as geometric ground truth. We specify data partitioning, sensor ablations, fault injection, paired closed loop testing, and real to simulation consistency assessment. The central hypothesis is that explicit evidence management can reduce dangerous false free space predictions and occlusion risk within a matched computational budget while retaining acceptable driving efficiency. All performance claims remain subject to experimental validation.

## 1 Introduction

An autonomous driving world model must support specific decisions: whether a vehicle can pass through a narrow gap, whether it should slow before an occluded region, and how other road users may affect candidate trajectories. Image reconstruction error and object detection accuracy alone do not answer these questions.

Tesla's public technical material emphasizes learning road structure and 3D objects from multiple camera streams, alongside continued development of end to end driving models and reinforcement learning [1, 2]. Waymo's recent public material emphasizes complementary camera, lidar, and imaging radar observations, together with learned sensor fusion [3, 4]. We draw on these public directions without assuming access to proprietary models, training data, or complete safety mechanisms. Historical demonstrations are not treated as specifications of current production systems.

Our research question is how video based semantics, direct geometric measurements, and generative scenario diversity can contribute to a unified system without conflating their evidential reliability.

We propose three design contributions for evaluation. First, we separate occupancy probability from evidence sufficiency so that unobserved space is not automatically interpreted as free. Second, we predict action conditioned futures over a layered 3D state and incorporate occlusion and tail risk into trajectory selection. Third, we establish a provenance aware reconstruction and generation loop whose final acceptance criterion is driving behavior consistency rather than rendering quality.

These contributions are methodological proposals. Their value relative to existing approaches must be established through ablations and closed loop testing.

# 2 Related Work and Scope

## 2.1 Occupancy and representations for planning

OccFormer studies vision based 3D semantic occupancy, including vertical structure beyond a planar bird's eye representation [5]. UniAD organizes multiple driving tasks within a planning oriented network [6]. We follow the principle that representations should serve decisions, with additional emphasis on evidence provenance, unknown space, and geometric checks relevant to safety. We do not claim that occupancy modeling or joint training is itself new.

## 2.2 Neural scene reconstruction

NeRF represents a continuous radiance field for novel view synthesis [7]. 3D Gaussian Splatting uses explicit Gaussian primitives for efficient radiance field rendering [8]. Street Gaussians separates urban backgrounds from moving vehicles [9]. These methods can support the creation of recorded road assets, but reconstructed densities or Gaussian distributions are not automatically physical surfaces suitable for collision checking.

## 2.3 Generative world models

Copilot4D studies generation of future lidar observations conditioned on ego motion [10]. The Waymo World Model publicly demonstrates camera and lidar generation with action, layout, and language controls [11]. We distinguish scene representation, future dynamics, and sensor generation. These components can be combined, but the ability to render a 3D scene does not establish reliable traffic interaction modeling.

# 3 Problem Formulation

At time t, the observation is defined as

$$Y_t = \{I_t, L_t, R_t, E_t\} \quad (1)$$

Here, I denotes multiple camera images, L lidar measurements, R imaging radar measurements, and E inertial, wheel speed, and vehicle state measurements. M denotes a map prior. Fusion operates over a time aligned and motion compensated history $Y_{t-H:t}$, retaining uncertainty in calibration and ego pose.

The unified world state is

$$S_t = \{G_t, O_t, A_t, T_t, V_t, U_t, Q_t\} \quad (2)$$

G describes road surfaces and clearance; O occupancy; A dynamic entities and track history; T road connectivity and traffic relations; V visibility and observation coverage; U uncertainty; and Q provenance, timestamps, and sensor health. The system estimates an approximate belief state $b_t = q_\theta(S_t \mid Y_{t-H:t}, \mathrm{M})$. This is an engineering approximation, not a claim that network outputs form an exactly calibrated posterior.

For each candidate ego action sequence $a^k$, the prediction model samples multiple possible future states:

$$S^{k,j}_{t+1:t+K} \sim p_\varphi(\cdot \mid b_t, a^k), \;\; j = 1, \dots, N \quad (3)$$

The objective is improved closed loop risk management subject to latency, memory, and power constraints. Perception accuracy, rendering fidelity, and forecasting error are diagnostic measures rather than sufficient evidence of system effectiveness.

## 4 Online 3D World Modeling

Figure 1 summarizes the online architecture and its offline learning loop. Current observations constrain the online world state, while reconstructed and generated scenes support training and evaluation under explicit evidence and provenance controls.

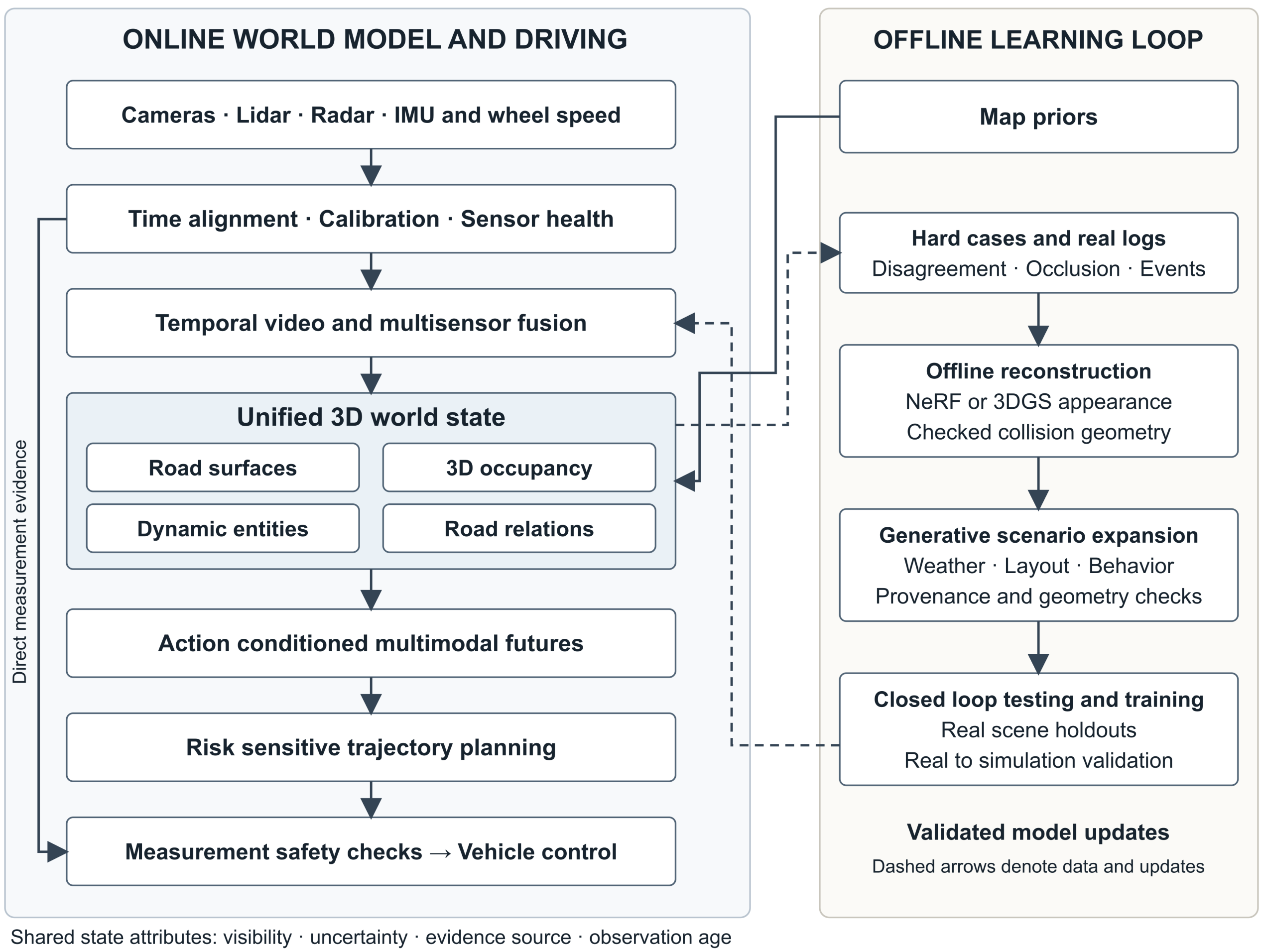

Figure 1. Overall DriveWorld3D architecture. Solid arrows show online processing and offline workflow; dashed arrows show hard-case transfer and model updates. Visibility, uncertainty, provenance, and observation age accompany the world state. Generated hypotheses support offline learning and do not replace current measurements.

### 4.1 Alignment sensor health and map validation

The system accounts for camera exposure timing, lidar scan timing, and radar measurement timing. Lidar observations are corrected for ego motion, and image projection uses available shutter information. Calibration residuals, timing drift, image obstruction, and abnormal returns contribute to a sensor health vector $h_t$.

Maps provide static road priors but cannot override persistent current obstacle evidence. When maps conflict with observations, the system reduces the relevant prior weight, records a map change event, and increases planning margins. Conflicts are retained before fusion so that feature aggregation does not make them untraceable.

### 4.2 Four layers of world state

Road surfaces. A local height field with sparse 3D structures describes slope, curbs, and depressions. Bridges, tunnels, and overpasses use multiple surfaces or local volumetric representations rather than a single height assumption. Traversability accounts for vehicle width, swept volume, ground clearance, and slope limits.

Volumetric occupancy. Block sparse voxels represent collision relevant space, with semantic, motion, and observation support attributes. The occupancy layer can retain irregular obstacles even when entity detection fails.

Dynamic entities. Object queries retain shape, velocity, covariance, and association history. Rigid vehicles use object coordinates, while pedestrians, cyclists, and deformable objects can use nonrigid occupancy residuals. Entity and occupancy outputs are cross checked without requiring every occupied region to belong to a recognized object.

Road relations. A graph represents lane connectivity, boundaries, traffic light associations, and right of way. Physical traversability and traffic permission are separate outputs so that empty space is not automatically a legal route.

### 4.3 Evidence and unknown space

For voxel i, the system predicts occupancy probability $p_i$ and evidence sufficiency $c_i$ separately. Unknown is an epistemic state, not a physical class equivalent to occupied. The discrete observation state is

$$d_i = \begin{cases} occupied & c_i \geq \tau_c, p_i \geq \tau_o \\ free & c_i \geq \tau_c, p_i \leq \tau_f \quad \tau_f < \tau_o \\ unknown & otherwise \end{cases} \quad (4)$$

Evidence sufficiency depends on sensor coverage, ray visibility, health, and observation age; it is not identical to classifier confidence. Historical support decays according to object class and motion state:

$$c_i^{prior}(t) = c_i(t_0)\, exp[-\lambda_i(t - t_0)] \quad (5)$$

Equation (5) controls expiration of old support and is not a complete Bayesian update. Unstable associations or uncertain motion require faster decay. Missing lidar returns do not unconditionally establish free space along an entire ray: valid range, weather, and return quality must be considered. Sensor evidence is correlated and cannot be accumulated indefinitely under an independence assumption.

### 4.4 Temporal fusion and measurement checks

The main network lifts video features into sparse 3D space and fuses them with lidar, radar, and temporal memory. A separate checking path retains valid measurement based near field obstacles and tests the vehicle's swept volume. Conflicts increase uncertainty and trigger more conservative planning.

This path reduces reliance on one fusion network, but shares some sensors with the main system and therefore has common cause failure modes. It is not a fully independent safety guarantee.

### 4.5 Risk driven spatial resolution

The scene is divided into computational blocks b. Proximity to candidate trajectories, time to potential collision, vulnerable road users, and uncertainty determine a priority score:

$$w_b = \alpha r_b + \beta u_b + \gamma v_b + \delta s_b \quad (6)$$

The normalized terms represent collision relevance, uncertainty, vulnerable user relevance, and surface or clearance sensitivity, respectively. Scores allocate blocks within a fixed budget. A minimum coverage budget prevents initially missed or low confidence targets from permanently losing computation. Training may use soft priorities, while deployment uses hierarchical selection with hysteresis to reduce resolution oscillation. At highway speeds, distant portions of the driving corridor can remain high priority.

## 5 Future Prediction and Risk Sensitive Planning

### 5.1 Action conditioned prediction

The predictor generates both multimodal entity trajectories and future occupancy. Near term forecasts emphasize motion continuity; longer horizons express interaction branches. Latent variable or diffusion models are possible implementations, but model size and sample count must respect onboard budgets.

Conditioning on an ego action does not demonstrate causal interaction learning. Logged actions are affected by selection bias. Counterfactual simulation and controlled tests must assess whether the model has learned useful responses rather than correlations alone.

### 5.2 Occlusion risk

Unknown regions are not uniformly treated as obstacles. Reachable risk regions are constructed using whether a plausible participant could enter the ego conflict region within the planning horizon and reasonable bounds on participant speed. Unknown space behind a bus near a crosswalk receives a different risk weight from space inside a closed building outside the road.

When probabilities are unreliable, bounded motion reachable sets support stress tests. Their costs are conservative risk proxies rather than precise collision probabilities.

### 5.3 Trajectory objective and safety filtering

For candidate trajectory $\tau$ and sampled scene cost $C_j(\tau)$, we use

$$\begin{aligned} J(\tau) &= \frac{1}{N} \sum_j C_j(\tau) + \eta\, CVaR_\alpha[C_j(\tau)] \\ &+ \lambda_u\, C_{unknown}(\tau) + \lambda_c\, C_{comfort}(\tau) + \lambda_r\, C_{rules}(\tau) \end{aligned} \quad (7)$$

Conditional value at risk emphasizes the cost of worse outcomes. Collision checks use expanded vehicle swept volumes and temporal occupancy, accounting for localization and geometry error. Spatial voxels are correlated; multiplying marginal voxel probabilities does not establish trajectory safety probability.

Trusted current obstacles, vehicle dynamics, and braking constraints first filter unacceptable candidates. Remaining trajectories are ranked by the objective. If none is feasible, the vehicle executes a predefined minimum risk strategy appropriate to its operating domain and capabilities. Finite samples and constraint checks do not establish formal safety in unrestricted environments.

# 6 Offline Reconstruction and Generative Training

## 6.1 Separate visual assets and collision geometry

Scenes are reconstructed from camera sequences, lidar, and offline optimized poses. Dynamic 3DGS is the preferred rendering representation; NeRF may be selected for particular appearance or view modeling requirements. Both are not required simultaneously. Static backgrounds and object assets are separated to support modified trajectories and layouts.

Collision geometry comes from checked surfaces, meshes, or occupancy volumes and is explicitly registered to appearance assets. Rendered pixels, geometric labels, and sensor outputs must refer to the same scene state. A Gaussian opacity threshold is not a universal physical surface definition.

## 6.2 Generative expansion and provenance

Scenario variables include illumination, weather, target position and motion, temporary road changes, and sensor faults. Each asset and label records one of three provenance categories: measurement supported, offline estimate supported, or generated hypothesis supported, together with its coverage and confidence.

Training rules depend on task and provenance. Generated scenes can support appearance robustness and stress tests, but unsupported regions are excluded from high confidence metric depth supervision. Generated variants remain grouped with their source scene to prevent train test leakage.

Generation checks cover ground contact, scale, interpenetration, road relations, and temporal continuity. Rain and snow testing must distinguish visual changes, sensor degradation, and tire road friction. Changing an image alone does not reproduce complete adverse weather conditions.

## 6.3 Training objectives and stages

The proposed total loss is

$$\begin{aligned}\mathcal{L} &= \lambda_o \mathcal{L}_{occ} + \lambda_g \mathcal{L}_{geom} + \lambda_a \mathcal{L}_{agent} + \lambda_t \mathcal{L}_{topo} \\ &+ \lambda_p \mathcal{L}_{pred} + \lambda_v \mathcal{L}_{visibility} + \lambda_m \mathcal{L}_{temporal} + \lambda_d \mathcal{L}_{drive}\end{aligned} \tag{8}$$

Occupancy loss is evaluated only where labels have valid support. Geometry loss uses robust distance residuals. Entity loss includes association and motion, while prediction loss uses an appropriate likelihood or generative objective. Visibility loss uses rays, occlusion, and offline coverage labels. Temporal loss excludes genuine scene changes. Driving loss combines demonstrations and closed loop objectives. Probability quality is evaluated with Brier score and negative log likelihood, with calibration on an independent validation set if needed.

Stage one trains geometry and temporal state. Stage two trains prediction and planning. Stage three jointly fine tunes the system while retaining geometric supervision, followed by closed loop hard case iteration. Loss weights, synthetic data proportions, and thresholds are selected on validation data, not test data.

## 6.4 Hard case selection

Sensor disagreement, intrusion into unknown regions, forecast disagreement, interventions, and abnormal braking identify upload candidates. Random samples are also retained to estimate the broader distribution.

Stress test results are reported separately after oversampling rare events; their sampled frequencies must not be interpreted as real road occurrence rates.

## 7 Experimental Protocol and Falsifiable Hypotheses

### 7.1 Research hypotheses

H1. Explicit evidence states reduce dangerous false free predictions in occluded scenes without an unacceptable reduction in driving efficiency.

H2. Under matched model size and runtime budgets, geometric constraints from multiple modalities improve risk management in low visibility scenes.

H3. Generated scenarios screened for geometry and provenance improve closed loop performance on held out real scenes more than an equal amount of unscreened generated data.

H4. Risk driven resolution improves geometry in critical regions at matched runtime cost without materially increasing missed obstacles.

Each hypothesis can be rejected. Unknown space modeling could merely increase unnecessary braking, and generated data could degrade performance on real scenes.

### 7.2 Data and partitions

Waymo Open Dataset or nuScenes can support multimodal perception evaluation, while nuPlan supports planning research [12]. Differences in sensor configurations, formats, and labels require separate training and evaluation. Metrics from different datasets cannot be combined as evidence for one integrated system. State level nuPlan testing does not validate sensor rendering; sensor level closed loop evaluation requires an additional environment with reactive agents.

Partitions are grouped by geography, acquisition time, and original log. Reconstructions, weather variants, and counterfactual variants of one log belong to the same partition. Separate subsets address in domain driving, unseen regions, low visibility, sensor degradation, and occlusion. Test data must include real observations that have not been generatively modified.

### 7.3 Baselines and ablations

| Configuration | Purpose |
|---|---|
| B0 Video temporal occupancy and a common planner | Vision baseline, not a Tesla reproduction |
| B1 Multimodal fusion without explicit evidence management | Measure the role of geometric measurements |
| B2 B1 plus visibility, provenance, and age decay | Evaluate unknown space management |
| B3 B2 plus action conditioned future prediction | Evaluate forecasting contribution |
| B4 B3 plus training with reconstructed real scenes | Evaluate reconstruction assets |
| B5 B4 plus screened generative expansion | Evaluate the complete proposal |

| Configuration | Purpose |
|---|---|
| B6 Equal generated data volume without provenance screening | Test generation quality controls |

Additional ablations remove lidar, radar, temporal memory, or adaptive resolution, and replace tail risk with mean risk. Missing modality evaluation includes both matched training and sudden deployment faults. Baselines match training data, optimization steps, and planner settings. Report both equal compute comparisons and results under each model's separately optimized budget.

### 7.4 Metrics and statistics

Perception metrics include occupancy intersection over union, obstacle recall within critical driving corridors, surface distance error, and unknown space identification. Dangerous false free rate is the fraction of reliably labeled occupied voxels in critical regions that the system declares confidently free. Free space precision, coverage, and risk coverage curves are also reported so that predicting unknown everywhere cannot produce a misleading advantage.

Primary closed loop metrics include collisions, near conflicts, task completion, route progress, unnecessary harsh braking, comfort, and traffic rule violations. Collision reporting distinguishes responsibility and scenario type, and includes simulation hours, distance, and raw event counts. Runtime reporting includes end to end latency at P50, P95, and P99, memory, power, and deadline misses, together with hardware and input dimensions.

Paired tests use the same initial state, traffic random seed, and route; multiple training seeds are reported. Confidence intervals resample independent scenarios or logs rather than correlated frames. Rare events use appropriate count or binomial intervals. Zero observed collisions do not imply zero risk. A pilot study and power analysis determine sample size.

### 7.5 Real to simulation consistency

Paired controlled track and simulation runs compare braking onset, minimum separation, steering, and event order. Evaluation also checks whether risk rankings across different driving policies are preserved. Replaying a recorded trajectory alone cannot establish counterfactual fidelity after the ego vehicle deviates from that trajectory. Report geometric rejection rates and rendering failures, not only successful examples.

### 7.6 Planned results reporting

| Method | False free rate | Collisions per simulation hour | Completion | Unnecessary harsh braking | P99 latency |
|---|---|---|---|---|---|
| B0 | Not evaluated | Not evaluated | Not evaluated | Not evaluated | Not evaluated |
| B1 | Not evaluated | Not evaluated | Not evaluated | Not evaluated | Not evaluated |
| B2 | Not evaluated | Not evaluated | Not evaluated | Not evaluated | Not evaluated |
| B5 | Not evaluated | Not evaluated | Not evaluated | Not evaluated | Not evaluated |

Table entries are placeholders for future experiments, not predictions or measured results. Lower is better except for completion. Operational definitions and thresholds must be fixed before the final test set is evaluated.

## 8 Limitations and Deployment Scope

Multiple sensors increase hardware, cleaning, calibration, and maintenance costs. Redundancy does not eliminate shared occlusions or extreme weather failures. Evidence thresholds and risk weights can cause excessive conservatism, requiring simultaneous efficiency evaluation. Uncertainty estimates can be miscalibrated, and measurement checks cannot detect every systematic error.

Generated scenes may violate physics. A world model and a driving model may share training biases, creating a validation loop that accepts their common mistakes. Independent real tests and controlled track validation therefore remain necessary. Online action conditioned forecasting is also limited by logged action coverage and computation. Initial validation targets urban roads within a defined operating design domain, rather than unrestricted roads and weather.

The proposed combination has not undergone an exhaustive novelty review. Architectural integration alone does not establish algorithmic priority, patentability, or superiority over recent commercial systems.

## 9 Conclusion

DriveWorld3D connects temporal video understanding, multimodal geometric evidence, unknown space management, and offline generative simulation in a 3D framework oriented toward driving decisions. The central design principle is to manage measured facts, historical estimates, and generated hypotheses separately, and to assess their value through closed loop risk, efficiency, and runtime cost. The next step is to implement budget matched baselines and the evidence state module, first testing dangerous false free predictions in occluded scenes, then isolating the contributions of future prediction and generative training. Until those experiments are completed, the conclusions concern system design and testability, not demonstrated performance or safety improvements.

## References

The following papers and official technical materials support the background discussion. Corporate publications describe publicly disclosed directions and are not independent performance evidence. Web materials accessed October 14, 2026.